\documentclass[runningheads]{llncs}
\usepackage[T1]{fontenc}
\usepackage{colortbl}
\usepackage{graphicx}
\usepackage[hidelinks]{hyperref}
\usepackage{multirow}
\usepackage{newfloat}
\usepackage{listings}
\usepackage{comment}
\usepackage{times}
\usepackage{multirow}
\usepackage{booktabs}
\usepackage{placeins}
\usepackage{amssymb}
\usepackage{amsmath}
\usepackage{orcidlink}
\newlength\myHeight 
\newlength\myWidth

\usepackage[acronym]{glossaries}

\newacronym{ai}{AI}{Artificial Intelligence}
\newacronym{ig}{IG}{Integrated Gradients}
\newacronym{gradcam}{GradCAM}{Gradient-weighted Class Activation Mapping}
\newacronym{knn}{k-NN}{k-Nearest Neighbors}
\newacronym{lrp}{LRP}{Layer-wise Relevance Propagation}
\newacronym{irof}{IROF}{Iterative Removal Of Features}
\newacronym{mst}{MST}{Minimum Spanning Tree}
\newacronym{mstc}{MST-C}{Minimum Spanning Tree Compactness}
\newacronym{roi}{ROI}{Region Of Interest}
\newacronym{rra}{RRA}{Relevance Rank Accuracy}
\newacronym{road}{ROAD}{Remove and Debias}

\newacronym{shap}{SHAP}{SHapley Additive exPlanations}
\newacronym{xai}{XAI}{Explainable Artificial Intelligence}

\title{Structural Compactness as a Complementary Criterion for Explanation Quality}
\titlerunning{Structural Compactness as a Criterion for XAI Quality}

\author{Mohammad Mahdi Mesgari \inst{1}\orcidID{0009-0008-9010-2073} \and 
Jackie Ma\inst{1}\orcidID{0000-0002-2268-1690}
Wojciech Samek\inst{1,2,3}\orcidID{0000-0002-6283-3265}
Sebastian Lapuschkin\inst{1,4}\orcidID{ 0000-0002-0762-7258} \and
Leander Weber\inst{1}\orcidID{0000-0003-2995-6914}}
\authorrunning{MM. Mesgari et al.}
\titlerunning{Structural Compactness as a Criterion for XAI Quality}
\institute{Fraunhofer Heinrich Hertz Institute, Berlin, Germany \and
Technische Universität Berlin, Berlin, Germany \and
BIFOLD – Berlin Institute for the Foundations of Learning and Data, Berlin, Germany \and
Centre of eXplainable Artificial Intelligence, Technological University Dublin, Dublin, Ireland
}

\begin{document}
\maketitle
\vspace{-0.5cm}

\begin{abstract}
In the evaluation of attribution quality, the quantitative assessment of explanation \emph{legibility} is particularly difficult, as human perception is influenced by the varying shapes and internal organization of attributions, which are not captured by simple statistics such as entropy, sparsity, or the number of salient values. To address this issue, we introduce \gls{mstc},  a \emph{graph-based structural metric} that combines higher-order geometric properties of attributions, such as \emph{spread} and \emph{cohesion}, into a measure of \emph{compactness}. Our method favors attributions with salient points spread across a small area and organized spatially into few cohesive clusters. To the best of our knowledge, it is the first metric to quantitatively capture these higer-order structural aspects of human perception, expressing numerically which heatmaps are perceived as legible by humans. We show that \gls{mstc} reliably distinguishes between explanation methods, exposes fundamental structural differences between models, and provides a robust, self-contained diagnostic for explanation compactness that complements existing notions of attribution complexity.
\end{abstract}

\section{Introduction}
Deep neural networks are increasingly deployed in high-stakes domains, where transparency, reliability, and trustworthiness are critical, driving a growing interest in explainability. Here, aside from methods that seek to understand the functionality of network components and their semantic meaning \cite{kim2018interpretability,dreyer2025mechanistic}, \emph{local} techniques such as feature attributions \cite{bach2015pixel,LundbergL17,SundararajanTY17,Selvaraju_2019} remain staple tools due to their simplicity, computational efficiency, and ability to localize relevant parts of the input and the model \cite{ross2018improving,weber2025efficient}. However, assessing the quality of attributions (encompassing both perturbation- and saliency-based techniques) remains a challenge~\cite{hedstroem2024fresh}, with numerous quantitative metrics having been proposed over the years to address this issue. 

While a large number of techniques focus on attribution \emph{correctness} (e.g., \cite{rieger2020irof,rong2022evaluating,Arras_2022,hedstroem2024fresh}, comprising categories such as localization, randomization, or faithfulness), a second, similarly relevant property is whether obtained attributions can be \emph{interpreted by humans}. Because querying human experts is costly, several works aim to instead build predictive \cite{Kazmierczak2024Pasta} or quantitative proxies for explanation legibility, readability, or understandability. Existing metrics mainly quantify explanation \emph{complexity} \cite{nguyen2020quantitative,BhattWM20,ChalasaniC00J20} through simple statistics, such as sparsity or the number of salient points. While these techniques capture basic properties of attribution value distribution, they overlook higher-order geometric characteristics, including structural organization. Consequently, they fail to model the topology of distinct salient regions, limiting not only their ability to fully characterize the spatial structure of explanations, but also their legibility, since explanations have been argued to be more interpretable when they form visually coherent regions \cite{bokadia2022evaluating} rather than fragmented pixel-level responses. Similarly, attributions with a low-complexity spatial distribution of salient regions seem to be favored by users \cite{Zhao2023Graphical}.

\begin{figure}[!ht]
    \centering
    \includegraphics[width=0.8\columnwidth]{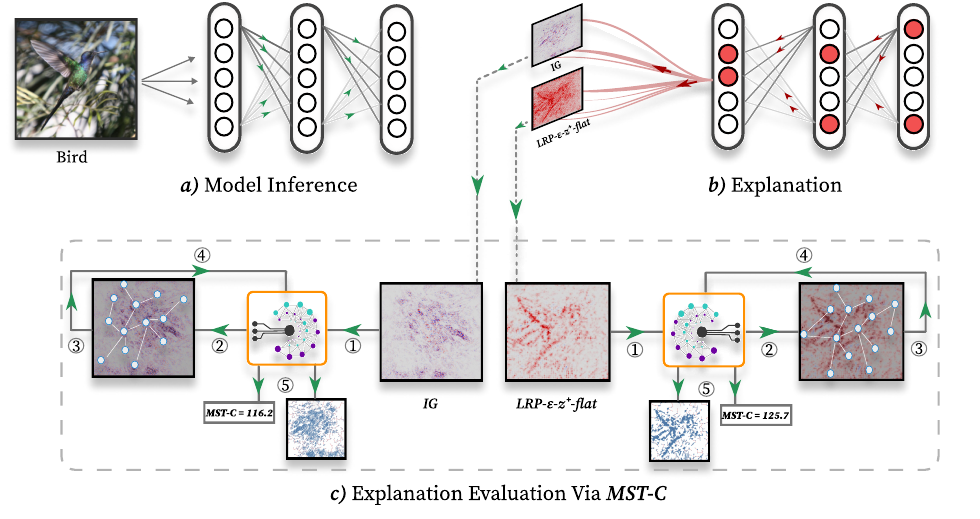}
    \caption{Overview of \gls{mstc} workflow. Given a sample, a prediction (a) and different attributions (b) are obtained. (c) \gls{mstc} scores are derived from each attribution, by (1) converting it into a graph, (2) obtaining its \gls{mst} preserving spatial structure (3) computing the \gls{mst} length, (4) capturing attribution \emph{spread} and \emph{cohesion}, and (5) aggregating these properties into a final \gls{mstc} score.}
    \label{fig:mstc-overview}
\end{figure}

To address these limitations, we introduce Minimum Spanning Tree Compactness (MST-C), a novel graph-based metric for quantifying attribution compactness in the image domain. Building upon existing notions of complexity \cite{nguyen2020quantitative,BhattWM20,ChalasaniC00J20}, \gls{mstc} jointly captures attribution \emph{spread} and \emph{cohesion} via Minimum Spanning Tree (MST) topology, augmented by convex hull area. As such, \gls{mstc} captures quantitative aspects of attribution \emph{legibility}, but is entirely unrelated to attribution \emph{correctness} or \emph{faithfulness}. We note that legibility is not equivalent to understandability, and rather its precursor~\cite{labarta2025cue}. An overview of the workflow for deriving \gls{mstc} is given in Figure~\ref{fig:mstc-overview}. Our experiments demonstrate that \gls{mstc} captures meaningful aspects of spatial organization, reveals structural differences between explanations, complements existing metrics that quantify explanation complexity, and remains robust across hyperparameter choices.
In summary, we make the following contributions:
\begin{itemize}
    \item We propose \gls{mstc}, a novel quantitative, graph-based metric for evaluating the structural \emph{compactness} of attribution-based explanations. \gls{mstc} combines attribution \emph{spread} and \emph{cohesion}, capturing higher-order geometric characteristics of attributions relating to their legibility.
    
    \item Through several examples, we demonstrate that \gls{mstc} not only captures visual differences in the qualitative legibility of attributions but also identifies model-specific attribution patterns.
    
    \item We compare \gls{mstc} to existing metrics for evaluating attribution complexity, showing that while it correlates with these metrics at a high level, it also provides additional, complementary insights by distinguishing attributions based on their structural organization.
    
    \item We evaluate the sensitivity of \gls{mstc} to its hyperparameters, provide recommendations, and investigate in-depth which properties cause attribution to exhibit high compactness as measured by \gls{mstc}.

\end{itemize}

\section{Methodology}\label{sec:methodology}

Given an RGB input image $x \in \mathbb{R}^{h\times w \times 3}$, a model $\mathcal{F}: \mathbb{R}^{h\times w \times 3} \rightarrow \mathbb{R}^c$, with $\mathcal{F} \in \mathbb{F}$ produces an output $\mathcal{F}(x) \in \mathbb{R}^c$ with $c$ elements. Here, $\mathbb{F}$ defines the space of possible models. Then, we define an attribution as a function $\mathcal{R}: \mathbb{R}^{h\times w \times 3} \times \mathbb{F} \rightarrow \mathbb{R}^{h\times w}$ that, based on a single input and the model, assigns importance scores to each input pixel. This definition encompasses both saliency-based (e.g., \cite{bach2015pixel,SundararajanTY17,Selvaraju_2019}) and perturbation-based (e.g., \cite{Fong2017Interpretable,Ribeiro2016Why,LundbergL17}) attributions. 

By representing attributions as graphs, \gls{mstc} captures the spatial organization of attribution maps and unifies two complementary aspects of compactness, \emph{Spread} and \emph{Cohesion} into a single score. Specifically, supported by previous research \cite{Zhao2023Graphical,Kazmierczak2024Pasta,labarta2025cue}, we argue that an attribution is more legible when the most salient points are spread across a smaller area and spatially organized into few but cohesive clusters.

\subsubsection{Graph Construction.}
\label{sec:graph-construction}

As an initial step, before converting the attribution heatmap into a graph, a thresholding operation is applied. Since both positive and negative attribution values can be important for explanation interpretation, the heatmap is first transformed by taking the absolute value of the attributions. The most salient pixels are then selected using thresholding based on attribution value percentiles. This step retains the most important pixels regardless of whether they contribute positively or negatively to the prediction, while filtering out low-value and noisy pixels. Because the goal is to analyze the spatial and structural properties of attribution maps, focusing on the most salient regions provides a more stable and meaningful representation.

After this preprocessing step, the attribution map is converted into a simple graph $G = (V, E)$, consisting of nodes $V$ and edges $E$. The selected salient pixels form the nodes, while edges $(v_i, v_j) \in E$ are weighted by the Euclidean distance $d(v_i, v_j)$ between nodes. The graph is constructed using \emph{\gls{knn}}, where each node is connected only to its $k$ nearest neighbors.

It is crucial that the constructed graph forms a single connected component, as the minimum spanning tree (\gls{mst}) requires connectivity to quantify attribution cohesion properly. Since using \gls{knn} does not guarantee full connectivity, we must choose $k$ large enough to ensure that all nodes belong to a single connected component. As detailed below, this graph representation allows for the quantification of both the attribution spread and cohesion.

\subsubsection{Quantifying Attribution Spread.}
Based on the attribution graph $G$, we first derive a notion of overall attribution spread as the area  $A_{\mathrm{hull}}$ of the convex hull enclosing all salient points. Since in the context of compactness, we are interested in the concentration of the spread over a small area, we take the inverse of $A_\text{hull}$ (specifically, its square root, for numerical stability) to define \emph{Attribution Spread} as 

\begin{equation}
q_\text{spread} = \frac1{\sqrt{A_{\mathrm{hull}}}}
\end{equation}

To make the measure more robust in outlier cases where there are not enough points to form a convex hull, i.e., $|V| < 3$, we set $\sqrt{A_{\mathrm{hull}}} = \sqrt{h^2 + w^2}$ (the attribution diagonal) where $h$ and $w$ are the height and width of the attribution map. Larger values of $q_\text{spread}$ indicate a tight spread of salient points, whereas smaller values correspond to high dispersion.

\subsubsection{Quantifying Attribution Cohesion.}
To quantify attribution cohesion, we first compute the \emph{\gls{mst}}~\cite{cormen2009introduction} of $G$, defined as the connected, acyclic subgraph, $T = (V, E_T)$, with $E_T \subseteq E$ that minimizes the total edge weight, yielding a simplified yet informative structure capturing the topology of salient attributions. We quantify this structural information via the length of the \gls{mst}, given by $L_T = \sum_{(v_i, v_j) \in E_T} d(v_i, v_j)$. The inverse of $L_T$, multiplied by the number of salient points $|V|$, then measures average structural coherence. We define the \gls{mst}-based \emph{Attribution Cohesion} as: 

\begin{equation}
q_\text{cohesion} = \frac{|V|}{L_T}
\end{equation}

Larger values of $q_\text{cohesion}$ indicate tightly clustered, well-connected points, whereas smaller values correspond to dispersed or fragmented points.

\subsubsection{Quantifying Attribution Compactness.}
In a final step, we combine $q_\text{spread}$ and $q_\text{cohesion}$ to directly define the \gls{mstc} metric as

\begin{equation}
\text{MST-C} = {q_\text{spread}} \cdot q_\text{cohesion}
\end{equation}

This multiplicative formulation allows \gls{mstc} to capture the extent to which salient pixels form compact or fragmented structures in relation to their overall spatial spread. 

\subsubsection{Theoretical Bounds of \gls{mstc}.}
We show that \gls{mstc} can be interpreted as an \emph{absolute} measurement of attribution compactness. To support this interpretation, we demonstrate that it can be theoretically bounded in the 2D image domain as follows:

Images are represented on a 2D regular lattice grid. Therefore, the minimum Euclidean distance between two distinct pixels in this space is $1$. Based on this property, we can derive bounds for both $q_{\text{spread}}$ and $q_{\text{cohesion}}$. For $q_{\text{spread}}$, the smallest possible convex hull area in the 2D lattice occurs when three graph nodes (i.e., salient pixels) form the smallest triangle, which has an area of $\tfrac{1}{2}$. The largest possible area corresponds to the convex hull covering the entire image grid with height $h$ and width $w$. Consequently, $q_{\text{spread}}$ can be bounded by

\begin{equation}
0 \leq \frac{1}{\sqrt{hw}} \leq \frac{1}{\sqrt{A_{\mathrm{hull}}}} \leq{\sqrt{2}} 
\end{equation}

Note, that even if nodes are colinear, we still define the convex hull to surround them as $h^2+w^2$.  With a minimum of $1$ given a single-pixel image, this edge-case also fulfills the above bound. 

For $q_{\text{cohesion}}$, the upper bound occurs when two nodes are separated by the minimum possible distance of $1$. This also implies that $q_{\text{cohesion}}$ must always be positive, bounding it by

\begin{equation}
0 \leq q_{\text{cohesion}} \leq 2
\end{equation}

\noindent Multiplication of the bounds of $q_{\text{cohesion}}$ and $q_{\text{spread}}$ yields an overall bound for \gls{mstc}: 

\begin{equation}
0 \leq q_{\text{spread}} \cdot q_{\text{cohesion}} \leq 2\sqrt{2}
\end{equation}

This bound holds for any configuration of pixels in the 2D image domain. As we found \gls{mstc} values to be typically small in practice, we scale the metric by a constant $C$ to improve numerical interpretability and facilitate clearer comparisons.
Considering the formulation of \gls{mstc} and empirical observations from our experiments, we set $C = \sqrt{h^2 + w^2}$, which provides well-scaled values and performs well in practice.

\section{Experiments}
\label{sec:experiments}
In the following Section, we evaluate \gls{mstc} across supervised and self-supervised models operating in the image domain and a diverse set of explainers (Section~\ref{sec:experiments:structure}). We then experimentally relate \gls{mstc} to metrics evaluating attribution complexity (Section~\ref{sec:experiments:complexity}) and finally investigate the sensitivity of our metric to hyperparameters (Section~\ref{sec:experiments:sensitivity}).
\paragraph{Dataset.}
We use the PASCAL VOC 2012 dataset \cite{pascal-voc-2012} as a primary benchmark. This dataset comprises 20 object classes across diverse real-world scenes, offering complex spatial structures suitable for attribution analysis. As a preprocessing step, all images are resized to $224 \times 224$ pixels and normalized using the standard ImageNet \cite{imagenet} channel-wise statistics.

\paragraph{Models.}
We evaluate \gls{mstc} using both supervised and self-supervised architectures: for the supervised setting, we use ResNet34 \cite{HeZRS16} and VGG-16 \cite{SimonyanZ14a} with pretrained ImageNet weights from \texttt{torchvision} \cite{torchvision}, and for the self-supervised setting, we employ an ImageNet pretrained SimCLR~\cite{abs-2002-05709} encoder from PyTorch Lightning Bolts \cite{plbolts}.

\paragraph{Attribution Methods.}
We apply several well-established explanation methods from different subcategories to generate attribution maps for evaluation. \emph{Gradient-based} approaches include \gls{gradcam} \cite{Selvaraju_2019} (for ResNet and VGG models) and \gls{ig} \cite{SundararajanTY17}. \emph{Modified Backpropagation} is represented by \gls{lrp} \cite{bach2015pixel}, where we employ a diverse set of best-practice composites: LRP-$\varepsilon$,  LRP-$\varepsilon$-$\gamma$-$z^B$, LRP-$\varepsilon$-$z^+$-$\text{flat}$, as well as LRP-$\varepsilon$-$z^+$-$k\text{flat}$, which extends the flat rule to the $k$ lowest layers of a model and serves for evaluation purposes of \gls{mstc} only. In terms of \emph{Occlusion-based} explanations, we use \gls{shap} \cite{LundbergL17}, specifically its approximation via GradientSHAP\footnote{Note that for SimCLR, we take gradients w.r.t. the $\ell_2$-norm of the embedding, reflecting the contribution of input features to the representation magnitude}. Before computation of quantitative scores, we normalize attributions by their maximum absolute value to maintain consistency.

\paragraph{\gls{mstc} Hyperparameters and Visualization.}
\gls{mstc} depends on two main hyperparameters: the number of neighbors \(k\) and the percentile threshold. For consistency across experiments and comparability of results, we set \(k = 500\) and use a percentile threshold of 80\% (cf. Section \ref{sec:recommendations}), picking the salient points in the 20\% highest value range. For attribution visualization, heatmaps are generated using a diverging colormap, where \emph{blue} denotes negative and \emph{red} denotes positive attribution values.  \gls{mst}-graphs are plotted 
on the original input spatial coordinates, with \emph{blue dots} representing nodes and \emph{red lines} edges in the \gls{mst}.

\subsection{Structural Analysis}
\label{sec:experiments:structure}

\begin{figure}[!t]
    \centering
    \includegraphics[width=0.9\columnwidth]{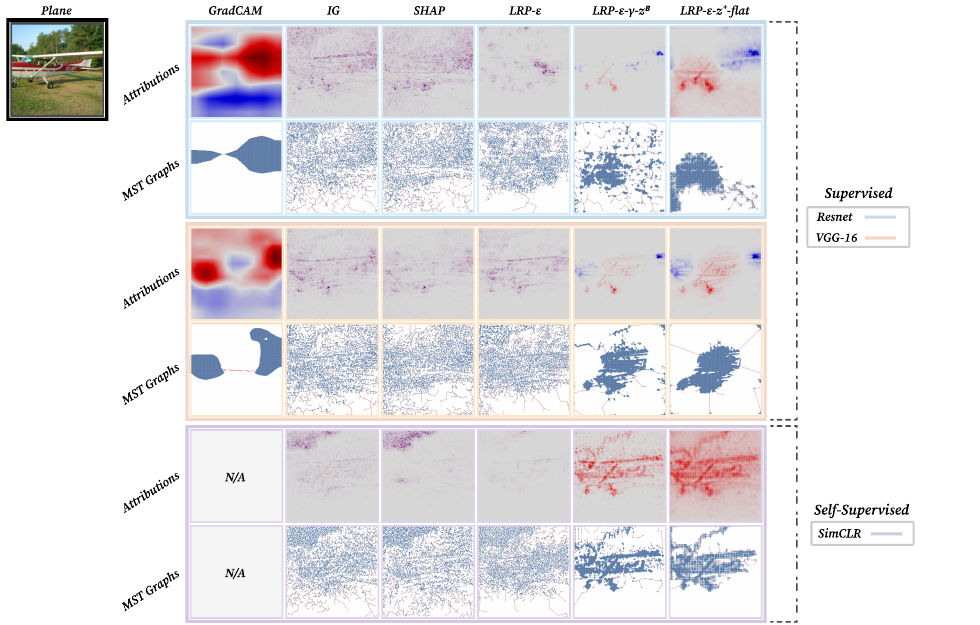}
    \caption{Attribution heatmaps (\emph{upper rows}) across \gls{xai} methods and models for a single sample, with corresponding \gls{mst} graph visualizations \emph{(bottom row)} highlighting structural differences. Refer to Figure \ref{fig:qual_comparison2} in the appendix for additional examples.
    }
    \label{fig:qual_comparison}
\end{figure}

To begin with, we use \gls{mstc} to quantitatively explore differences among the mentioned attribution methods in terms of their structural properties and investigate how the \gls{mstc} score relates to qualitative differences in their respective heatmaps across models and methods. Beyond the overall \gls{mstc} score, we also examine its main components: the \gls{mst} length, $L_T$, and the convex hull area, $A_{\mathrm{hull}}$. Due to the constant percentile threshold, the number of nodes $|V|$ also remains constant across our experiments. Example heatmaps for the different attribution methods and models are shown in Figure~\ref{fig:qual_comparison}, accompanied by visualizations of the corresponding \gls{mst}. Quantitative results are reported in Table~\ref{tab:xai_scores}.

\begin{table}[h!]
\centering
\caption{Comparison of \gls{mstc} and its main components across attribution methods and models. From left to right, we report the mean \gls{mstc} score (higher is better), mean \gls{mst} length $L_T$ (lower is better), and mean convex hull area $A_\text{hull}$ (lower is better), together with the respective standard deviations, computed across 250 samples. For each \gls{xai} method, we \underline{underline} the best values across models. For each model, we mark the best values across \gls{xai} methods in \textbf{bold}.}
\begin{tabular}{llccc}
\toprule
\textbf{XAI Method} & \textbf{Model} & $\overline{\text{\gls{mstc}}}\uparrow$ & \textbf{$\overline{L_{T}}\downarrow$} & \textbf{$\overline{A_{\mathrm{hull}}}\downarrow$} \\
\midrule
\multirow{2}{*}{\gls{gradcam}} 
 & ResNet & \textbf{\underline{265.03}} $\pm$ 48.33 & \textbf{10051.90} $\pm$ 3065.65& \textbf{\underline{123.99}} $\pm$ 46.45\\ 
    & VGG-16 & \textbf{197.82} $\pm$ 38.62 & 10129.17  $\pm$ 895.06 & 163.99 $\pm$ 30.03\\ 
\midrule
\multirow{3}{*}{Integrated Gradients} 
    & ResNet & 116.03 $\pm$ 4.13 & 12463.19 $\pm$ 224.04 & 220.07 $\pm$ 0.99\\ 
    & VGG-16 & 118.16 $\pm$ 7.12 & \underline{12281.07} $\pm$ 397.23 & 219.67 $\pm$ 0.95 \\ 
    & SimCLR & \underline{119.22} $\pm$ 6.77 & 12364.08 $\pm$ 308.59 & \underline{216.20} $\pm$ 1.13\\ 
\midrule
\multirow{3}{*}{SHAP} 
    & ResNet & 116.31 $\pm$ 5.77 & 12469.56 $\pm$ 224.57 & 219.59 $\pm$ 0.98 \\ 
    & VGG-16 & 118.00 $\pm$ 6.95 & 12287.91 $\pm$ 433.89 & 219.80 $\pm$ 1.61 \\ 
    & SimCLR & \underline{122.02} $\pm$ 8.82 & \underline{12168.85} $\pm$ 354.40 & \underline{214.93} $\pm$ 2.66 \\ 
\midrule
\multirow{3}{*}{LRP-$\varepsilon$} 
    & ResNet & \underline{131.30} $\pm$ 10.90 & 11561.32 $\pm$ 275.46 & \underline{210.40} $\pm$ 5.57 \\ 
    & VGG-16 & 119.09 $\pm$ 5.37 & \underline{12184.38} $\pm$ 333.78 & 219.44 $\pm$ 1.35 \\ 
    & SimCLR & 120.10 $\pm$ 7.64 & 12246.01 $\pm$ 319.62 & 216.80  $\pm$ 3.74\\ 
\midrule
\multirow{3}{*}{LRP-$\varepsilon$-$\gamma$-$z^B$} 
    & ResNet & 130.97 $\pm$ 3.33 & 11059.68  $\pm$ 175.68 & 219.61 $\pm$ 4.21 \\ 
    & VGG-16 & \underline{135.19} $\pm$ 2.38 & \underline{10579.65} $\pm$ 132.02 & 222.35 $\pm$ 16.93\\ 
    & SimCLR & 129.53 $\pm$ 6.40 & 11287.09 $\pm$ 234.40 & \underline{217.84} $\pm$ 7.06 \\ 
\midrule
\multirow{3}{*}{LRP-$\varepsilon$-$z^+$-$\text{flat}$} 
    & ResNet & \underline{144.66} $\pm$ 12.15 & 10872.84 $\pm$ 254.37 & \underline{203.34} $\pm$ 20.95 \\ 
    & VGG-16 & 139.99 $\pm$ 5.72 & \underline{10401.22} $\pm$ 97.33 & 218.66 $\pm$ 21.34\\ 
    & SimCLR & \textbf{135.51} $\pm$ 10.25 & 11462.11 $\pm$ 275.03 & 205.58 $\pm$ 10.45\\ 
\bottomrule
\end{tabular}
\label{tab:xai_scores}
\end{table}

\subsubsection{Attribution Methods: Comparison and Interpretation.}
We first focus on differences between attribution methods, and how they impact the \gls{mstc} score and its components. As shown in Figure~\ref{fig:qual_comparison}, methods such as \mbox{\gls{lrp}-$\varepsilon$}, \gls{ig} and \gls{shap} produce pixel-level attributions that contain a high degree of detail but can be difficult to interpret. Especially between \gls{ig} and \gls{shap}, our analysis reveals a striking topological similarity (cf. graph visualizations in Figure~\ref{fig:qual_comparison}), which may be rooted in their shared theoretical foundation as Aumann–Shapley cost-sharing frameworks \cite{SundararajanN20}. The quantitative \gls{mstc} scores (Table \ref{tab:xai_scores}) confirm this assertion. Notably, these visually noisy methods produce the lowest, almost identical \gls {mstc} scores (together with \mbox{\gls{lrp}-$\varepsilon$}). Nevertheless, \gls{mstc} uncovers nuanced morphological divergences, with \gls{shap} exhibiting higher variance and a consistently smaller $A_{\mathrm{hull}}$. The stochastic sampling in the utilized \gls{shap} implementation may introduce inherent variability, leading to attribution clusters that exhibit higher average compactness but larger inter-sample variations.

Despite also producing similarly fine-grained attributions, the \gls{mst} plot for \mbox{\gls{lrp}-$\varepsilon$} (Figure~\ref{fig:qual_comparison}) differs visually for ResNet, fragmenting attribution patterns into spatially isolated clusters. Consequently, \gls{mstc} is comparatively large for \mbox{\gls{lrp}-$\varepsilon$} (Table \ref{tab:xai_scores}) and ResNet. Closer inspection of the heatmaps in Figure~\ref{fig:qual_comparison} suggests that the discrepancy observed in ResNet arises from a downsampling artifact caused by the shortcut connections \cite{weber2023beyond} artificially inflating structural density and consequently \gls{mstc} scores. For VGG-16 and SimCLR models, \mbox{\gls{lrp}-$\varepsilon$} behaves similarly to \gls{ig} and \gls{shap} in terms of compactness, both visually and quantitatively.

In contrast, \mbox{\gls{lrp}-$\varepsilon$-$\gamma$-$z^B$} and \mbox{\gls{lrp}-$\varepsilon$-$z^+$-flat} composites concentrate relevance around a coherent “core” of supportive features. Again, this is reflected in \gls{mstc} scores in Table \ref{tab:xai_scores}, where especially \mbox{\gls{lrp}-$\varepsilon$-$z^+$-flat} attains a large mean score but also has a large standard deviation in comparison to \mbox{\gls{lrp}-$\varepsilon$-$\gamma$-$z^B$}. This behavior may arise from the Flat rule in the first layer, where relevance is redistributed uniformly across the receptive field rather than guided by learned weights, acting as a spatial aggregator and producing denser attributions, as reflected by the reduced $L_{T}$. However, the elevated standard deviation indicates that this increased compactness comes at the cost of stability.

To complete our comparison of attribution methods, we finally consider \gls{gradcam}, which produces coarse, feature-level attributions by upsampling a low-level attribution map through large parts of the model. Topologically, this creates a distinct signature with visually clear, focused, and high-density regions (cf. Figure~\ref{fig:qual_comparison}). Correspondingly, $A_{\mathrm{hull}}$ is comparatively small and $L_{T}$ significantly shorter than for other methods, resulting in the highest \gls{mstc} scores. However, this bias toward smoothness comes at the cost of increased variability; the inherent coarse resolution (typically $7 \times 7$) may cause disproportionate spatial shifts that depend on the model's focus on high-level features, rendering the topology more sensitive to global feature saliency rather than local pixel-wise evidence.

\subsubsection{Model Architectures: Comparison and Interpretation}
Next, we examine the effect of model architecture on attribution topologies and the resulting \gls{mstc} scores. As reported in  Table~\ref{tab:xai_scores}, \gls{mstc} scores for \gls{ig} and \gls{shap} remain largely consistent across all model types, with variations below 5\%, indicating minimal backbone dependency for these methods. Minor differences in $L_T$ and $A_{\mathrm{hull}}$ reflect subtle variations in spatial dispersion; for example, SimCLR’s higher \gls{mstc} for \gls{shap} ($122.02 \pm 8.82$) suggests more interconnectedness, while its lower $A_{\text{hull}}$ ($214.93$) indicates that attribution values remain spatially localized.

\mbox{LRP-$\varepsilon$} shows a slightly stronger sensitivity to model architecture, largely due to the ResNet outlier values caused by the downsampling artifact. In contrast,  \mbox{\gls{lrp}-$\varepsilon$-$\gamma$-$z^B$} and \mbox{\gls{lrp}-$\varepsilon$-$z^+$-flat} produce relatively similar structural properties. For both composites, VGG-16 attains the lowest $L_T$, indicating a highly compact attribution structure. For \mbox{\gls{lrp}-$\varepsilon$-$z^+$-flat}, ResNet exhibits a dense spatial distribution, possibly due to a minor effect of the observed downsampling artifact.

\gls{gradcam} highlights backbone-dependent patterns in feature map attributions. Its \gls{mstc} scores vary substantially, with ResNet exhibiting the highest relevance concentration ($265.03$) and the smallest convex hull area ($A_{\text{hull}} = 123.99$). This behavior reflects how deep residual connectivity and high-level feature pooling result in small, dense attributions. However, the substantially higher standard deviation for ResNet ($\sigma = 48.33$) compared to VGG-16 ($\sigma = 38.62$) indicates that, although ResNet’s attributions are more concentrated on average, they exhibit considerably greater variability across inputs.

\subsection{Comparison with Other Metrics of Attribution Quality}
\label{sec:experiments:complexity}
\begin{figure}[!t]
    \centering
    \includegraphics[width=0.8\columnwidth]{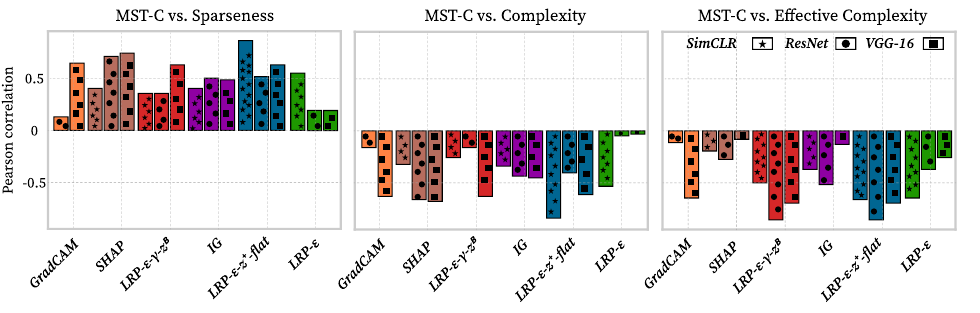}
    \caption{Correlation of \gls{mstc} with Sparseness, Complexity, and Effective Complexity across 250 samples, spanning different attribution methods and models. As shown, \gls{mstc} exhibits a strong positive correlation with Sparseness (up to ~0.85 for LRP-$\varepsilon$-$z^+$-$\text{flat}$, on the SimCLR architecture) and negative correlations with Complexity and Effective Complexity.}
    \label{fig:correlation_complexity}
\end{figure}

To situate \gls{mstc} within the broader XAI evaluation landscape, we compare it against representative metrics, specifically from the \emph{complexity} category: Since \gls{mstc} aims to capture attribution compactness, a structural property influencing legibility of explanations, it is fundamentally related to complexity-based metrics such as Sparseness \cite{ChalasaniC00J20}, Complexity \cite{BhattWM20}, and Effective Complexity \cite{nguyen2020quantitative}, as implemented in Quantus~\cite{HedstromWKBMSLH23}. We investigate the Pearson Correlation Coefficient of \gls{mstc} to the above metrics, highlighting its role as a complementary diagnostic tool, that, while related, captures structural nuances not covered by global statistics. The result of this comparison is shown in Figure~\ref{fig:correlation_complexity}.

\begin{figure}[!t]
    \centering
    \includegraphics[width=0.8\columnwidth]{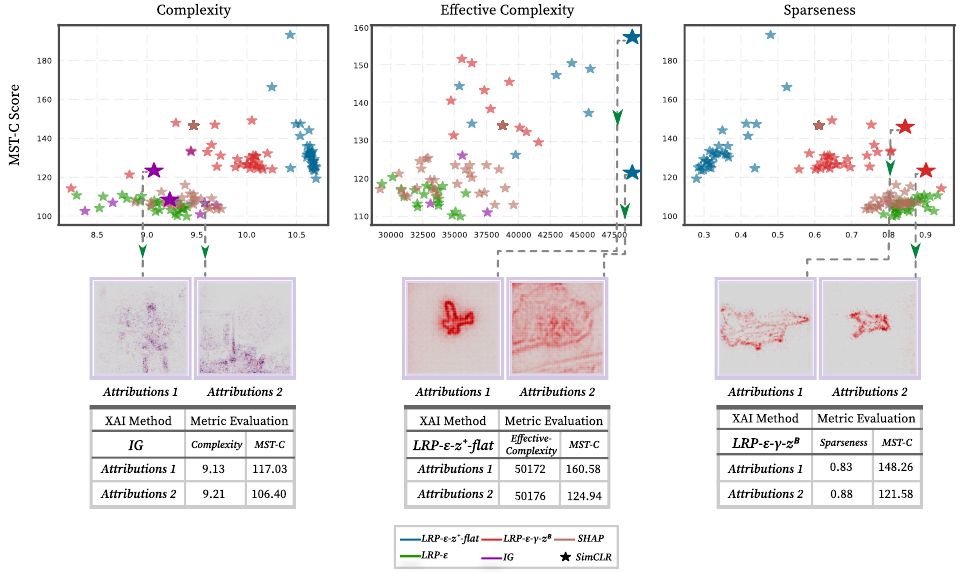}
    \caption{Comparison of samples with similar complexity metric scores but divergent \gls{mstc} values on the SimCLR architecture. While the samples receive comparable evaluations under Complexity, Effective Complexity, and Sparsity, they differ substantially in \gls{mstc}. The dashed lines highlight pairs of samples that obtain similar scores under each complexity metric but exhibit noticeably different spatial structures. This indicates that \gls{mstc} can reveal nuanced structural (and visually meaningful) differences in attribution maps that are not captured by traditional complexity metrics.}
    \label{fig:close_complexity_divergent_mstc}
\end{figure}

Here, we observe significant trends when comparing \gls{mstc} to complexity metrics. As depicted in Figure~\ref{fig:correlation_complexity}, \gls{mstc} exhibits a robust positive correlation with Sparseness, and a negative correlation with Complexity and Effective Complexity across all investigated architectures (note that with Complexity and Effective Complexity, lower is better, while with \gls{mstc} compactness and Sparseness, higher is better). Attribution maps with high Complexity and Effective Complexity scores are often characterized by a uniform and noise-like distribution of attribution values. This may correlate to the presence of fragmented regions where importance is scattered across the input space and thus relate to a lower \gls{mstc} score signifying reduced spatial coherence. Conversely, low Complexity may correlate to spatially concentrated attibutions that yield higher \gls{mstc} values. 
Crucially, \gls{mstc} demonstrates a unique ability to distinguish attributions that share similar complexity metric scores but exhibit different spatial topologies (cf. Figure~\ref{fig:close_complexity_divergent_mstc}). For example, despite a similar Effective Complexity score ($50172$ and $50176$), the shown \mbox{\gls{lrp}-$\varepsilon$-$z^+$-flat} attributions (Figure~\ref{fig:close_complexity_divergent_mstc} \emph{middle}) differ not only visually, but are also assigned vastly different \gls{mstc} values ($160.58$ and $124.94$), revealing structural differences that are not captured by Effective Complexity. A similar effect can be observed in the other shown examples, to a lesser degree. Together, these results suggest that \gls{mstc} provides a structural perspective that complements existing metrics based on global statistics by capturing structural nuances of spatial organization.

\begin{figure}[t]
    \centering
    \includegraphics[width=0.9\columnwidth]{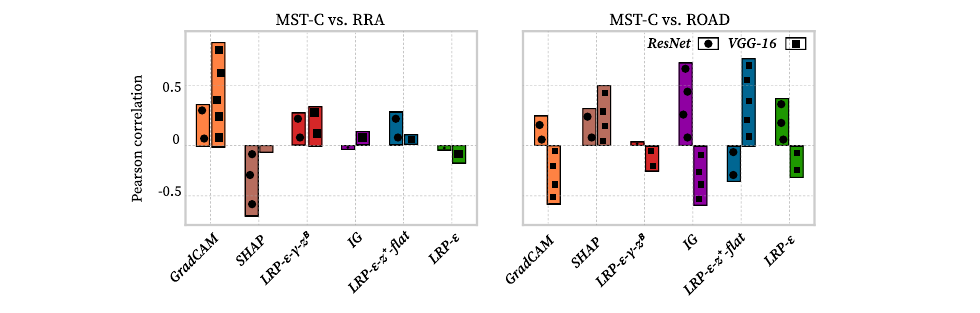}
    \caption{Correlation of \gls{mstc} with \gls{rra} \cite{Arras_2022} (\emph{left}) and \gls{road} \cite{rong2022evaluating} (\emph{right}) across 250 samples, encompassing multiple attribution methods and models. As expected, no consistent correlation is observed between \gls{mstc} and metrics evaluating aspects of attribution \emph{correctness}, such as localization (\gls{rra}) and faithfulness \gls{road}. In combination with results in Figure \ref{fig:correlation_complexity}, this underscores how \gls{mstc} captures aspects of attribution \emph{legibility}, independent of attribution correctness.}
    \label{fig:MST_C_vs_RRA_ROAD}
\end{figure}

For completeness, we also compare \gls{mstc} with \gls{rra}~\cite{Arras_2022} and the ROAD~\cite{rong2022evaluating} metric, representing the \emph{localization} and \emph{faithfulness} categories, respectively (cf. Figure \ref{fig:MST_C_vs_RRA_ROAD}). When compared to \gls{rra}, \gls{mstc} shows no significant numerical correlation. This lack of linear dependency underscores the complementary nature of legibility and correctness: an explanation can be structurally dense and coherent (high \gls{mstc}) while still being poorly aligned with the ground-truth regions of interest (low \gls{rra}) or unfaithful to the model (low \gls{road}). This observation highlights that structural "goodness" is not a proxy for correctness, emphasizing their orthogonal perspectives on explanation quality. 

\subsection{Hyperparameter Robustness of \gls{mstc}}
\label{sec:experiments:sensitivity}

Most quantitative \gls{xai} evaluation techniques are subject to several hyperparameters which the resulting scores can be highly sensitive to \cite{hedstroem2023metaeval}. For this reason, we assess the reliability of the obtained \gls{mstc} scores under different hyperparameter configurations and recommend best practice values.

\subsubsection{Robustness against Number of Neighbors k.}
\begin{figure}[t]
    \centering
    \includegraphics[width=0.8\columnwidth]{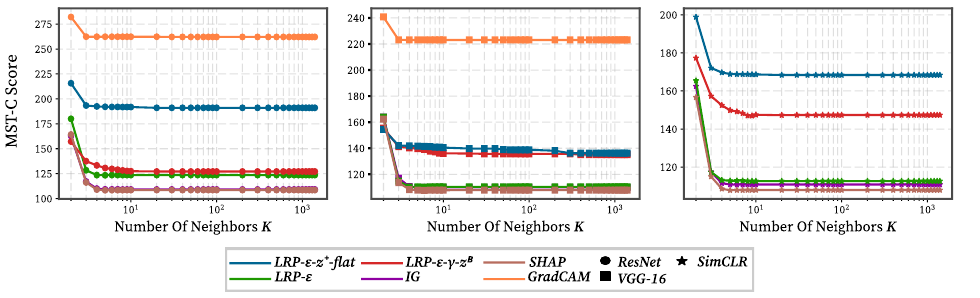}
    \caption{Average \gls{mstc} scores for different choices of graph construction hyperparameter $k$. Scores are shown for several attribution methods for SimCLR (\emph{left}), ResNet (\emph{middle}), and VGG-16 (\emph{right}), and avaraged across 250 samples. $k$ is varied between $0$ and $1500$. With a sufficiently large number of neighbors $k$, the \gls{mstc} scores stabilize, indicating convergence with respect to neighborhood size.}
    \label{fig:graph_construction_robustness}
\end{figure}

As discussed in Section~\ref{sec:graph-construction}, \gls{mstc} requires a graph representation of the attributions, where we employ \gls{knn} for graph construction. In Figure~\ref{fig:graph_construction_robustness}, we vary the number of neighbors $k$, in order to investigate the sensitivity of the obtained \gls{mstc} score against hyperparameter $k$. Notably, $k$ affects the cost of computing the \gls{mst}, with large $k$ vastly increasing computation time (worst-case $\mathcal{O}(k|V| \log(|V|))$), but small $k$ potentially leading to graph fragmentation. As shown in Figure~\ref{fig:graph_construction_robustness}, \gls{mstc} scores start stabilizing from $k=5$ upward across models and attributions, suggesting robustness to the choice of graph type and neighborhood size.

\subsubsection{Robustness against Attribution Thresholding.}
\begin{figure}[t]
    \centering
    \includegraphics[width=0.8\columnwidth]{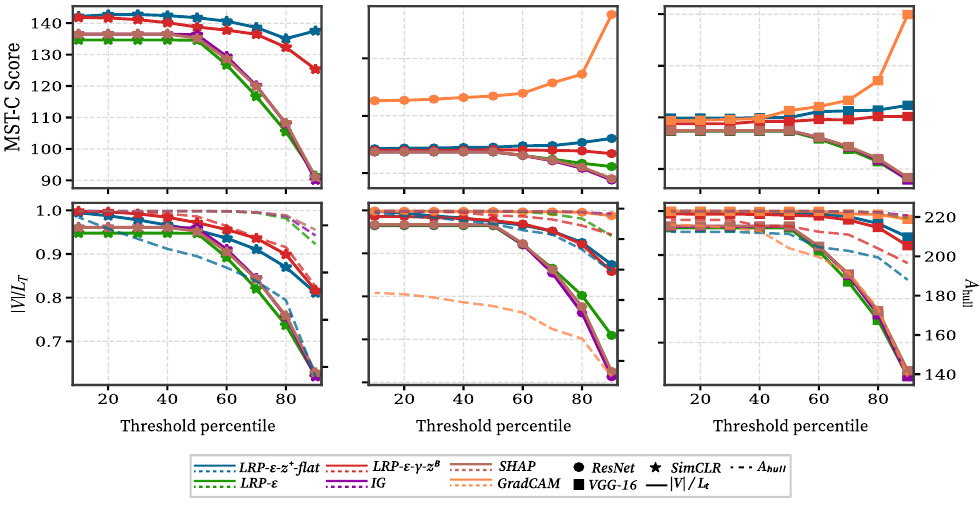}
    \caption{Sensitivity of \gls{mstc} (\emph{top}) and its components (\emph{bottom}) to threshold variation. The percentile-based threshold is varied between 10\% and 90\%. Especially at higher thresholds, \gls{mstc} scores are increasingly affected, with some variation in the size of this effect between attribution methods. Despite this, the ranking of attributions under \gls{mstc} remains relatively stable across thresholds.}
    \label{fig:sensitivity_threshold}
\end{figure}

The graph construction step of \gls{mstc} requires selection of salient attributions, as discussed in Section~\ref{sec:graph-construction}. For this purpose, we use a percentage threshold to essentially binarize attributions. In the following, we evaluate the sensitivity of \gls{mstc} to this threshold value, varying it from 0.1 to 0.9 (with higher values indicating higher selectiveness and less attribution values becoming graph nodes). The results of this experiment are visualized in Figure~\ref{fig:sensitivity_threshold}. 

At low thresholds, \gls{mstc} remains relatively stable, but increasingly diverges as thresholds become stricter. As visualized in Figure~\ref{fig:sensitivity_threshold} (\emph{bottom}), both \emph{cohesion} ($|V|/L_T$) and \emph{spread} ($A_{\mathrm{hull}}$) decrease as the threshold increases, resulting in progressively smaller and sparser graph representations. While this downward trend in raw components is consistent, different explainers exhibit slightly different trends. Fine-grained methods such as \mbox{\gls{lrp}-$\varepsilon$}, \gls{shap}, and \gls{ig} show a fast decline across metric components and consequently \gls{mstc}. For \gls{lrp}-$\varepsilon$-$\gamma$-$z^B$ and \mbox{\gls{lrp}-$\varepsilon$-$z^+$-flat}, the same effect can be observed, but to a lower degree (except for \mbox{\gls{lrp}-$\varepsilon$-$z^+$-flat} and ResNet). A fast increase in \gls{mstc} score can be observed for \gls{gradcam} on ResNet and VGG-16, and \mbox{\gls{lrp}-$\varepsilon$-$z^+$-flat} on ResNet. A fast decrease in the convex hull compared to cohesion creates an imbalance that causes a fast increase in overall \gls{mstc} score. Consequently, \gls{mstc} seems particularly sensitive to the thresholding in these cases. We note, however, that \gls{mstc} remains relatively stable for all attributions except \gls{gradcam}, and despite this divergence, the ranking of attributions under \gls{mstc} is not affected significantly.

\begin{figure}[t]
    \centering
    \includegraphics[width=0.8\columnwidth]{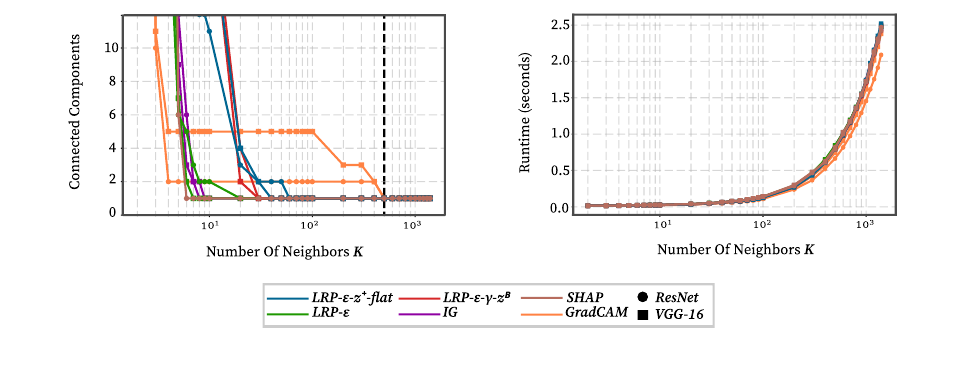}
    \caption{Number of connected components is shown in the \emph{left} panel, averaged over 250 samples. For $k = 500$ (illustrated as a black dashed line), all graphs form a single connected component. The \emph{right} panel shows the runtime (seconds per iteration), which increases with $k$, but remains reasonably fast at $k = 500$ (approx. 2.5 seconds).}
    \label{fig:graph_construction_further_robustness}
\end{figure}

\subsubsection{Practical Recommendations.}
\label{sec:recommendations}
While the optimal hyperparameters depend on the specific setting, our analysis leads to the following practical recommendations. Figure~\ref{fig:graph_construction_robustness} shows that the scores start converging around $k = 5$, but further investigation (Figure~\ref{fig:graph_construction_further_robustness}) across the tested samples suggests that $k \approx 500$ works well in practice to ensure that all constructed graphs form a single connected component. Performance monitoring confirms that this choice maintains single connected components across samples while keeping runtime reasonably fast (cf. Figure~\ref{fig:graph_construction_further_robustness}). Similarly, setting a percentile threshold of 80\% effectively filters noise, distinguishes between methods, and preserves the most salient attributions. For these reasons, we adopted this configuration consistently across the experiments in Sections~\ref{sec:experiments:structure} and \ref{sec:experiments:complexity}.

\subsection{Sensitivity to Attribution Resolution}

In our previous experiments, we observed \gls{gradcam} to perform particularly well under \gls{mstc} (cf. Table \ref{tab:xai_scores}), while also exhibiting unique behaviors under hyperparameter variation (cf. Figure~\ref{fig:sensitivity_threshold}). These observations are sensible, since \gls{gradcam} offers low-complexity and easily readable attributions due to its inherent coarseness. This unique property is caused by \gls{gradcam} computing low-resolution explanations for an intermediate convolutional feature map, which is then simply upscaled to input resolution. We nevertheless investigate in the following how this property affects \gls{mstc} scores. While in practice, attributions are usually compared at the same resolution, our analysis may offer insight into how the abstraction of fine-grained, lower-layer model computations can improve attribution compactness and possibly legibility.

\subsubsection{Sensitivity to Interpolation-based Upscaling of Attributions.} 
\begin{figure}[!t]
    \centering
    \includegraphics[width=0.8\columnwidth]{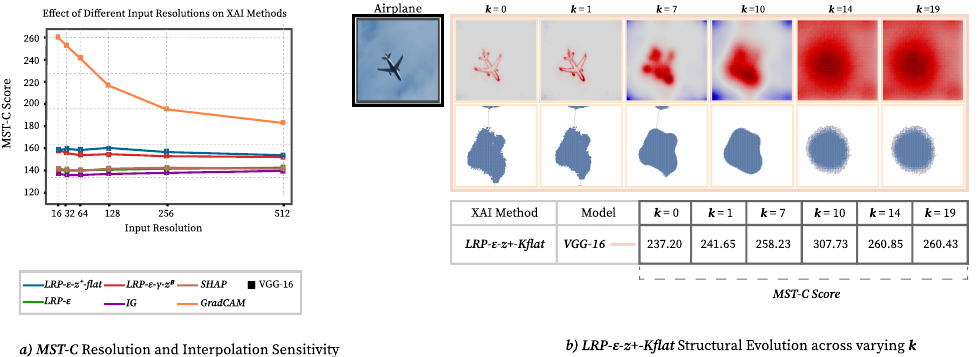}
    \caption{Effect of bilinear interpolation for VGG-16 per attribution method (\emph{left}) averaged over 250 samples. Effect of varying $k$ in the LRP-$\varepsilon$-$z^+$-$k\text{flat}$ composite (\emph{right}), via example heatmaps and graph plots \emph{(top)} and averaged \gls{mstc} scores \emph{(bottom)}.}
    \label{fig:sensitivity_Interp_Recep}
\end{figure}

For this purpose, we first investigate how simple bilinear interpolation (ranging from 16×16 to 512×512) affects the compactness of attribution heatmaps. As shown in Figure~\ref{fig:sensitivity_Interp_Recep} (\emph{left}), upscaling or downscaling attributions barely affects relative rankings between methods, but reduces the gap between the highest and lowest compactness attributions. Specifically, coarser methods (\gls{gradcam}, \mbox{\gls{lrp}-$\varepsilon$-$z^+$-flat}, and \mbox{\gls{lrp}-$\varepsilon$-$\gamma$-$z^B$}) exhibit high compactness at low resulutions, which decreases with higher resolutions, although not sufficiently to reach the low compactness of fine-grained techniques (\gls{shap}, \gls{ig}, and \mbox{\gls{lrp}-$\varepsilon$}). This observation is especially pronounced for \gls{gradcam} (as well as \mbox{\gls{lrp}-$\varepsilon$-$z^+$-flat} to a lesser degree), supporting the above hypothesis that deliberate omission of lower-layer computations may aid in improving attribution compactness by shifting focus to semantically meaningful structures rather than pixel-wise scores.

\subsubsection{Sensitivity to Receptive-Field-based Upscaling of Attributions.}  
We further investigate the effect of lower-layer abstraction through a variation of \gls{lrp} employing model-dependent upscaling: the LRP-$\varepsilon$-$z^+$-$k\text{flat}$ composite, which leverages the \emph{Flat} rule for the first $k$ layers of the model to upscale intermediate attributions based on neuron receptive fields. As shown in Figure~\ref{fig:sensitivity_Interp_Recep} (\emph{right}), heatmaps become visually denser with increasing $k$. Interestingly, the quantitative \gls{mstc} score reaches its maximum at $k = 10$ (307.73), exceeding the largest value reached by \gls{gradcam} in Figure~\ref{fig:sensitivity_Interp_Recep} (\emph{left}). Beyond this point, \gls{mstc} declines again as attributions cover a larger area while not increasing in cohesion, as visible in the ``fuzzy border'' of the corresponding \gls{mst} visualizations. We note that \gls{mstc} seems to have a bias towards blob-like attributions --- as the optimum found by \gls{mstc} at $k=10$ does not necessarily correspond to the visual optimum (around $k=1$). Despite this, it is able to distinguish both too finegrained ($k=1$) and too coarse ($k\geq14$) attributions. In terms of our above hypothesis, the abstraction of fine-grained, lower-layer model computations seems to positively affect attribution compactness up to a point, but result in a detrimental effect after that.

\section{Conclusion}
In this paper, we presented \gls{mstc}, a novel graph-based metric for the quantitative evaluation of attribution compactness. By representing attribution maps as spatial graphs and deriving two complementary properties, \emph{spread} and \emph{cohesion}, our metric captures structural aspects of human perception beyond global summarizing statistics such as entropy. I.e., \gls{mstc} numerically distinguishes which attributions are perceived as legible or not. Our results demonstrate that \gls{mstc} captures the structural properties of attributions, as reflected visually in their heatmaps, and uncovers legibility-related differences between attribution methods and models. Within the broader context of quantitative attribution evaluation, \gls{mstc} complements existing complexity metrics by distinguishing visual differences in attributions with similar complexity scores. Finally, we observed that abstraction of fine-grained lower-layer computations may improve attribution compactness.

Nevertheless, our method is subject to several limitations. Firstly, it depends on two hyperparameters, $k$ and the thresholding percentile. We investigate the effect of these hyperparameters in detail in Section~\ref{sec:experiments:sensitivity}, demonstrating consistency in relative attribution rankings under \gls{mstc} across wide hyperparameter ranges and providing practical value recommendations. Secondly, the thresholding effectively binarizes attributions, akin to \cite{nguyen2020quantitative}, thereby preventing our metric from distinguishing fine-grained value differences.
Thirdly, while \gls{mstc} scores and qualitative assessment of attribution legibility align, our metric seems to favor blob-like attributions. Even though these types of attributions may be blurry and not easily understandable, we argue that they are generally highly legible due to their large, connected areas and high contrast.
Furthermore, while our evaluation relies on comparisons with other quantitative complexity metrics and qualitative examples to assess the relationship between attribution legibility and \gls{mstc}, a user study directly validating the correlation between our metric and perceived explanation quality remains an important avenue for future work. 
Lastly, our formulation of \gls{mstc} and experimental analyses focus entirely on attributions in the image domain. While adaptation of our framework to other domains is possible in principle, provided that attributions are representable as graphs, this would require significant methodological adaptation, which we leave to future work.

\newcommand*{\img}[1]{%
    \raisebox{-.02\baselineskip}{%
        \includegraphics[
        height=\baselineskip,
        width=\baselineskip,
        keepaspectratio,
        ]{#1}%
    }%
}
\begin{credits}
\subsubsection*{\ackname} This work was supported by the European Union’s Horizon Europe research and innovation programme (EU Horizon Europe \img{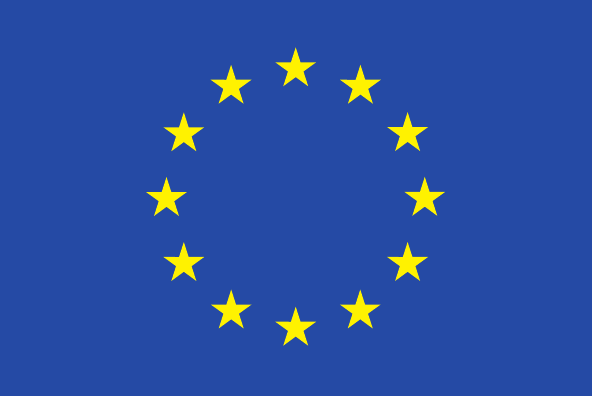}) as grants [ACHILLES (101189689), TEMA (101093003)]. This work was further supported by the Federal Ministry of Research, Technology and Space (BMFTR) as grant BIFOLD (01IS18025A, 01IS180371I); and the German Research Foundation (DFG) as research unit DeSBi [KI-FOR 5363] (459422098).

\subsubsection*{\discintname}
The authors have no competing interests to declare that are
relevant to the content of this article. 
\end{credits}

\bibliographystyle{splncs04}
\bibliography{references}

\renewcommand\thesubsection{A.\arabic{subsection}}
\renewcommand\thesection{}
\renewcommand\theequation{A.\arabic{equation}}
\renewcommand\thefigure{A.\arabic{figure}}
\setcounter{equation}{0}
\setcounter{figure}{0}
\appendix

\section{Additional Figures}
\begin{figure}[t]
    \centering
    \includegraphics[width=0.9\columnwidth]{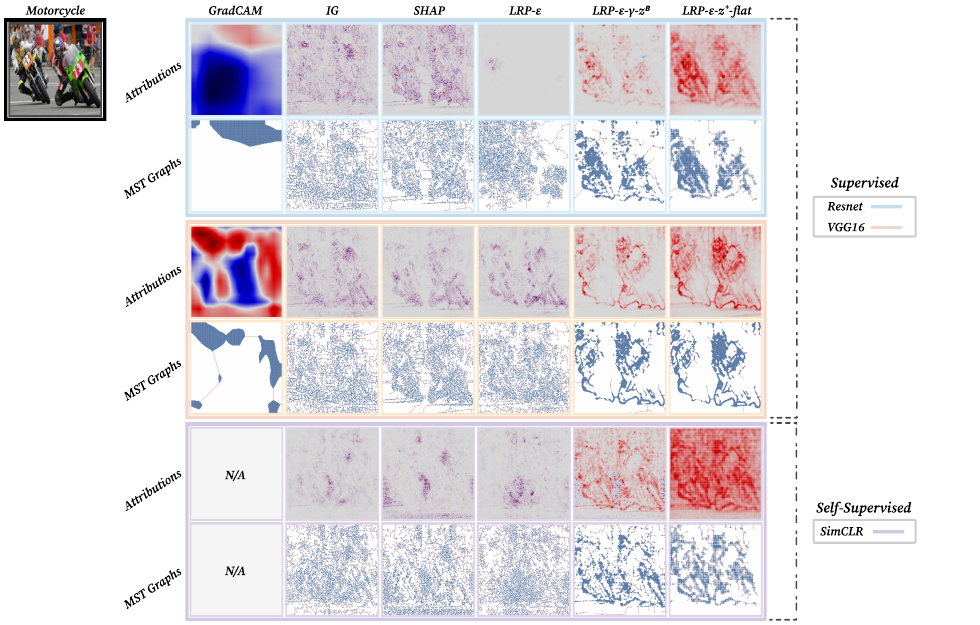}
    \includegraphics[width=0.9\columnwidth]{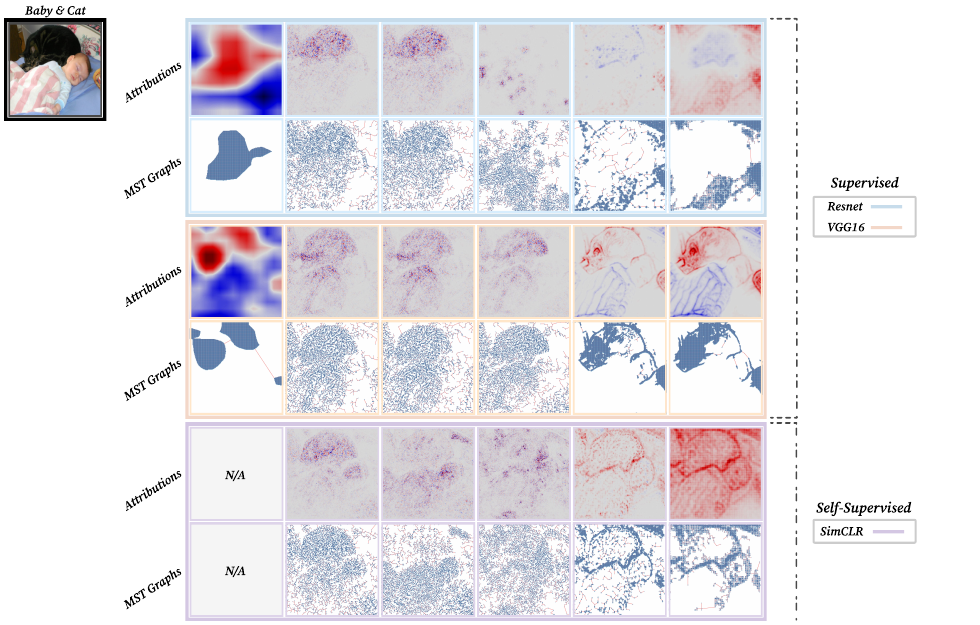}
    \caption{Additional examples, cf. Figure \ref{fig:qual_comparison}.}
    \label{fig:qual_comparison2}
\end{figure}

\end{document}